\documentclass[runningheads]{llncs}
\usepackage{graphicx}
\usepackage{subfiles}
\usepackage{booktabs}
\usepackage{multirow}
\usepackage{caption}
\usepackage{latexsym}
\usepackage {marvosym}
\usepackage{amsmath}
\usepackage[colorlinks,linkcolor=blue]{hyperref}

\begin {document}

\title{PE-MED: Prompt Enhancement for Interactive Medical Image Segmentation}

\titlerunning{PE-MED}

\author{Ao Chang\inst{1,2,3,4}\thanks{Ao Chang and Xing Tao contribute equally to this work.}, Xing Tao\inst{1,2,3\star} \and Xin Yang\inst{1,2,3} \and Yuhao Huang\inst{1,2,3} \and \\
Xinrui Zhou\inst{1,2,3} \and Jiajun Zeng\inst{1,2,3} \and Ruobing Huang\inst{1,2,3} \and Dong Ni\inst{1,2,3}\textsuperscript{(\Letter)}}

\authorrunning{Chang et al.}

\institute{\textsuperscript{$1$}National-Regional Key Technology Engineering Laboratory for Medical Ultrasound, School of Biomedical Engineering, Health Science Center, Shenzhen University, China\\
\email{nidong@szu.edu.cn} \\
\textsuperscript{$2$}Medical Ultrasound Image Computing (MUSIC) Lab, Shenzhen University, China\\
\textsuperscript{$3$}Marshall Laboratory of Biomedical Engineering, Shenzhen University, China\\
\textsuperscript{$4$}Shenzhen RayShape Medical Technology Co., Ltd, China\\}
\maketitle   
\begin{abstract}
Interactive medical image segmentation refers to the accurate segmentation of the target of interest through interaction (e.g., click) between the user and the image.
It has been widely studied in recent years as it is less dependent on abundant annotated data and more flexible than fully automated segmentation. 
However, current studies have not fully explored user-provided prompt information (e.g., points), including the knowledge mined in one interaction, and the relationship between multiple interactions.
Thus, in this paper, we introduce a novel framework equipped with prompt enhancement, called PE-MED, for interactive medical image segmentation.
First, we introduce a Self-Loop strategy to generate warm initial segmentation results based on the first prompt.
It can prevent the highly unfavorable scenarios, such as encountering a blank mask as the initial input after the first interaction.
Second, we propose a novel Prompt Attention Learning Module (PALM) to mine useful prompt information in one interaction, enhancing the responsiveness of the network to user clicks.
Last, we build a Time Series Information Propagation (TSIP) mechanism to extract the temporal relationships between multiple interactions and increase the model stability.
Comparative experiments with other state-of-the-art (SOTA) medical image segmentation algorithms show that our method exhibits better segmentation accuracy and stability. 
\keywords{Interactive Segmentation \and Prompt Learning.}
\end{abstract}

\section{Introduction}
Medical image segmentation is a pivotal aspect of research in medical image analysis, intended for the extraction of specific targets or regions in medical images for further analysis and diagnosis \cite{zhou2021review}. 
Traditional segmentation methods heavily rely on image processing techniques and machine learning algorithms, which can be computationally intensive, time-consuming, and require a high level of expertise. Deep learning-based methods have achieved state-of-the-art (SOTA) performance in automatic segmentation of medical images \cite{shen2017deep}. 
However, most current automatic methods lack the learning of informative prompts, resulting in inaccurate and inflexible segmentation. 

Interactive segmentation methods offer a promising solution to these challenges, utilizing limited user guidance to extract the target object and combining this with image features to yield finely segmented results~\cite{zhou2020embracing}. While user interaction leads to more precise segmentation results, the interaction process should be efficient and time-saving to reduce the burden on the user.

\textbf{Traditional interactive methods} for image segmentation use low-level features like edge information or color distribution, such as GraphCuts and Random Walk~\cite{hu2019fully,boykov2001interactive,criminisi2008geos,grady2006random}. 
These methods often require multiple user interactions and are time-consuming to produce satisfactory results as low-level features may not always distinguish the desired object from the background. 
To reduce user interactions and improve segmentation accuracy, machine learning techniques are leveraged. 
For instance, GrabCut \cite{rother2004grabcut} uses a Gaussian mixture model to estimate foreground and background distributions.
The initial results can be obtained via a user-provided bounding box, and refined through additional interactions.

Recently, \textbf{deep learning-based methods} have achieved SOTA performance in medical image segmentation~\cite{chen2021transunet,swinunet,heidari2023hiformer,huang2021flip}, thanks to the neural networks for automatically capturing high-level semantic features~\cite{shen2017deep}. 
Thus, deep models have been proposed to integrate with interactive methods for medical image segmentation.
There are two main streams of current approaches, as introduced below:

\textbf{-The first type of method ignores prompt learning during training}~\cite{bredell2018iterative,wang2018deepigeos,wang2018interactive,liu2022transforming}.
These methods require pre-training a semantic segmentation network, followed by fine-tuning the predicted mask through user interactions.
However, they may not be suitable for multi-class segmentation tasks that require accurate delineation of different targets.
Besides, they cannot ensure the quality of feature extraction when dealing with unfamiliar data patterns or categories, causing a performance drop.

\textbf{-The second type of method involves prompt learning during training.} 
Prompts are leveraged to guide the learning of deep models in these methods, including iSegFormer~\cite{liu2022isegformer}, Segment Anything Model~(SAM) \cite{kirillov2023segment,huang2023segment}, etc.
During testing, users need to interactively click on the foreground or background to achieve an accurate target segmentation.
Such prompt learning techniques have the potential to make interactive segmentation more flexible, accurate, and general to complex scenarios.
However, it is still challenging to deeply mine sparse prompts information and improve network response to user-provided prompts.

In this study, we propose a novel interactive approach with prompt enhancement to improve medical image segmentation performance, named PE-MED. Our contribution is three-fold. 
First, we employ a simple yet effective self-loop method to address the issue of insufficient information during the first interaction.
Second, we propose a novel Prompt Attention Learning Module (PALM) that explores the relationship between user interactions and image features to extract essential interaction details, enhancing the network's response to user input.
Third, we introduce a Time Series Information Propagation (TSIP) mechanism to model the continuity between multiple interactions for improving stability. 
Extensive experiments validated that, compared with the SOTA methods, our PE-MED can achieve accurate results with less user interaction.

\begin{figure}[!t]
    \centering
    \includegraphics[width=1\textwidth]{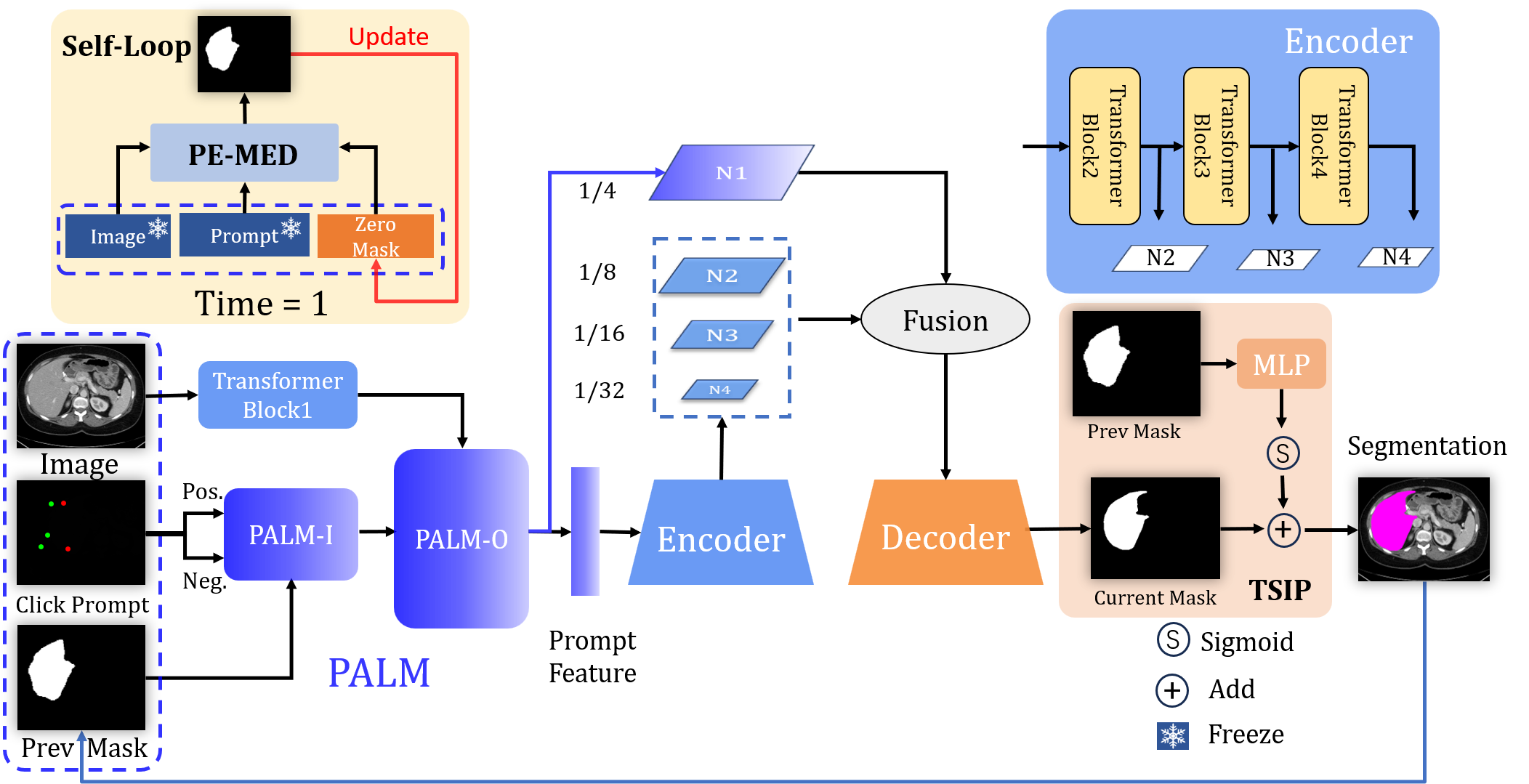}
     \caption{Overview of our proposed interactive segmentation framework. The green and red dots in the Click Prompt represent positive and negative prompt points, respectively. Sequential processing through four Transformer blocks (block1-4) enables the generation of multi-level features at $\{1/2^n,n=2,3,4,5\}$ of the original image resolution. The fusion module is composed of a convolutional layer and a normalization layer. The decoder module is a simple multilayer perceptron.}
    \label{fig:framework}
\end{figure}

\section{Methodology}

Fig.~\ref{fig:framework} shows the schematic view of our proposed method.
We propose the PE-MED, a novel iterative refinement framework equipped with prompt-enhanced modules, for medical image segmentation.
PE-MED consists of two stages:
1) The initial stage (see \textit{Time = 1}) serves as the foundation for the subsequent clicks,
2) The main stage represents the subsequent clicks after the initial one.
In the first stage, we first introduce a Self-Loop strategy to obtain a good initialized mask.
Then, the original image, previous mask and prompts (positive and negative points) are taken as inputs, and transmitted to the PALM and TSIP for prompt enhancement.
Last, the network will output the refined segmentation iteratively.

\subsection{Self-Loop Strategy for Warm Start}

For interactive segmentation methods, the initial segmentation generated by the user's first click plays a fundamental role in the subsequent interactions. Most of the current method lacks the optimization of the first interaction, resulting in poor initialization with inadequate prompt information. This will have a negative impact on the subsequent network module optimization and even segmentation failure. To address this issue, inspired by~\cite{zhou2020automatic,zhou2021multi}, we introduce a Self-Loop strategy to obtain a warm start for iterative segmentation (left-upper yellow block in Fig.~\ref{fig:framework}). Specifically, after the user's first click, fully empty masks are sent to the network, obtaining the rough segmentation (\textit{M0}). Then, the coarse prediction will enter the loop, and output the information-enhanced mask (\textit{M1}) without any interaction. Compared to \textit{M0}, \textit{M1} contains richer prompt information, and makes the learning of subsequent modules easier.

\subsection{PALM for Prompt Feature Enhancement}

\begin{figure}[!t]
    \centering
    \includegraphics[width=0.9\textwidth]{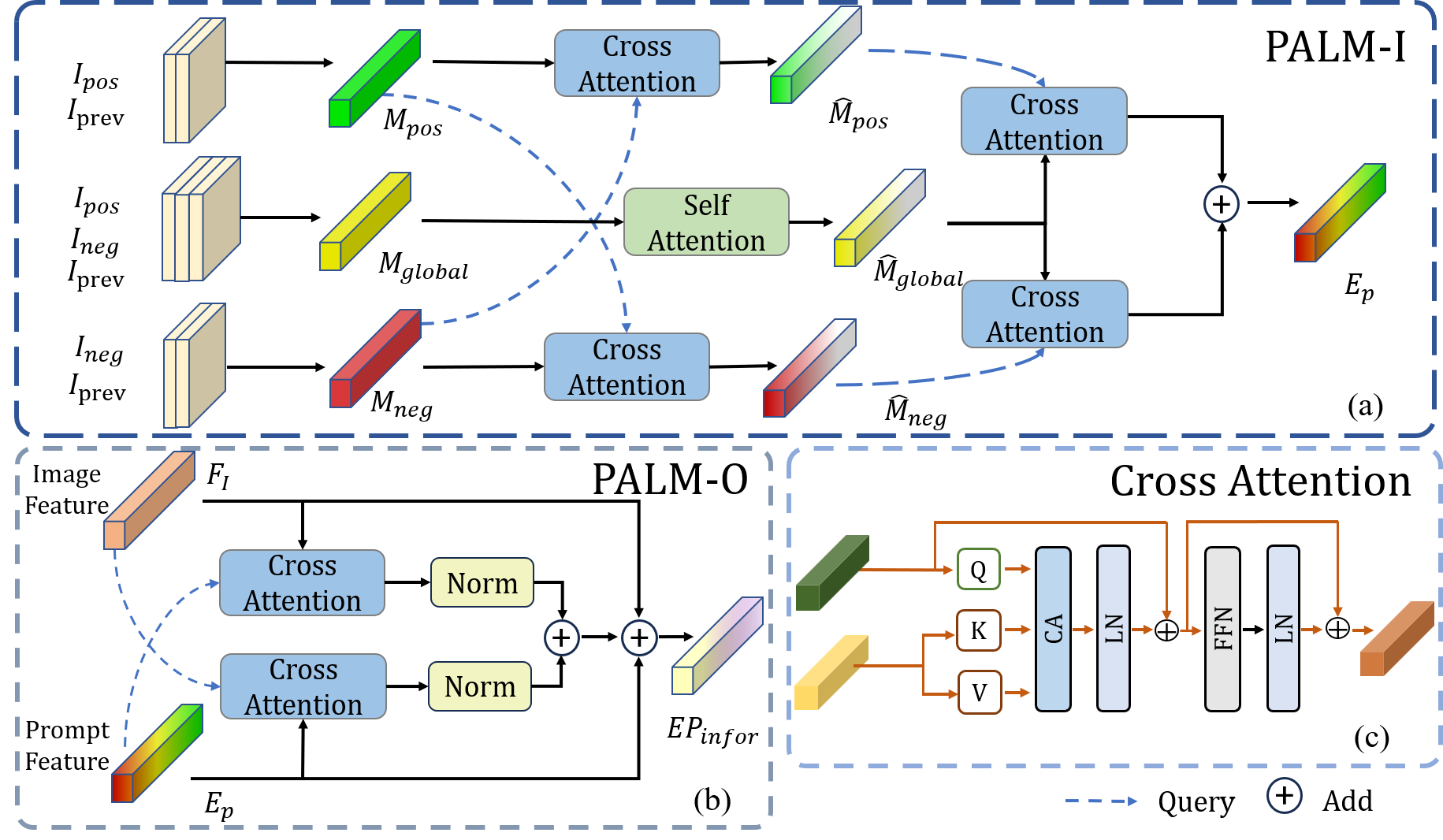}
    \caption{Overview of our proposed PALM.}
    \label{fig:method}
\end{figure}

Extracting rich information from sparse user hints is a challenging task.
In our study, we developed PALM to effectively leverage and enhance the prompt information.
As shown in Fig.~\ref{fig:method}, PALM consists of two modules: PALM-I (see Fig.~\ref{fig:method}(a)) and PALM-O (see Fig.~\ref{fig:method}(b)).
Specifically, PALM-I primarily focuses on augmenting the intrinsic features of prompts, whereas PALM-O enhances the interplay between prompts and images.

PALM takes four inputs, including the image $I_{image}$, positive and negative prompts ($I_{pos}$ and $I_{neg}$), and previous mask ($I_{prev}$). 
After the users input is converted to a disk map following a click, $I_{pos}$ and $I_{neg}$ are encoded and utilized in the PALM module.
As shown in Fig.~\ref{fig:method}(a), in the PALM-I part, we combine $I_{pos}$, $I_{neg}$, and $I_{prev}$ to form the concatenated feature map $M_{pos}$, $M_{global}$, and $M_{neg}$ using patch embeddings.
The Cross Attention Module (Fig.~\ref{fig:method}(c)) is then applied to enhance these feature maps, resulting in augmented features $\hat{M}_{pos} = \theta_{c} (M_{neg}, M_{pos},M_{pos})$ and $\hat{M}_{neg} = \theta_{c} (M_{pos}, M_{neg},M_{neg})$, with $\theta_{c}(Q,K,V) = Softmax(QK^T/d_k)V$. 
Simultaneously, the Self Attention Module improves $M_{global}$, yielding the global feature map $\hat{M}_{global}$:

\begin{equation}
\label{equa:eq3} 
    \hat{M}_{global} = \theta_{s} (M_{global}) = \theta_{c}(M_{global},M_{global},M_{global}) ,
\end{equation}

After obtaining enriched interaction and global information, we compute the prompt feature $E_{p}$ by summing the features resulting from cross-attentive mechanisms using $\hat{M}_{pos}$ and $\hat{M}_{neg}$ as query vectors, $E_{p}$ can be defined as:
\begin{equation}
    \label{equa:eq4} 
    \begin{aligned} 
    E_{p} = \theta_{c} (\hat{M}_{pos}, \hat{M}_{global}, \hat{M}_{global}) + \theta_{c}(\hat{M}_{neg}, \hat{M}_{global}, \hat{M}_{global}) ,
    \end{aligned} 
\end{equation}

Besides, we propose the PALM-O to bridge the information gap between prompt and image features. 
It takes $E_{p}$ and the image features $F_{I}$ (after Transformer Block 1) as input. 
Then, with the normalized layer ($norm$), the enhanced mixture feature $EP_{infor}$ can be illustrated as:
\begin{equation}
    \label{equa:eq5} 
    \begin{split}
    EP_{infor} = F_{I}  + norm(\theta_{c} (F_{I}, E_{p} , E_{p})) \\ + E_{p} + norm(\theta_{c} (E_{p} ,F_{I}, F_{I})).
    \end{split}
\end{equation}

\subsection{TSIP Mechanism for Stability Enhancement among Prompts}

Most of the existing interactive segmentation algorithms often ignore the relationship between consecutive interactions. 
It may lead to poor segmentation stability, especially when dealing with multiple interactions. 
To overcome this limitation, we propose to integrate TSIP mechanism into the framework to enhance the stability among multiple prompts.

The TSIP mechanism enables the extraction of dynamic and continuous interaction information, with the entire network functioning as a cohesive unit to convey temporal information. 
Specifically, the previous network output serves as a candidate memory for the current moment using a simple multilayer perceptron (MLP), while the current output serves as a candidate memory for the next moment. This mechanism is mathematically represented by:

\begin{equation}
\label{equa:eq6} 
    \begin{aligned}
        &O_{t} = F(I_{input}) + Sigmoid(\theta(O_{t-1})) ,
    \end{aligned}
\end{equation}
where $F$ refers to PE-MED without TSIP, $\theta$ refers a MLP. $I_{input}$ denotes the input image, click prompts and previous mask, $O_{t-1}$ and $O_{t}$ represent the output of the previous and current network, respectively.

\section{Experimental Results}

\begin{table}[t!]
    \caption{Method comparison on \textbf{Synapse} dataset. \textit{-p*} denotes the number of point prompts.}
    \centering
        \resizebox{\textwidth}{!}
    {
        \begin{tabular}{c|c|cccccccccccccc}
        \toprule  
            \multirow{2}{*}{Methods}& 
            \multicolumn{9}{c}{DSC(\%) $\uparrow$} \\
             \cline{2-10} 
            & AVG & Aorta & Gallbladder & Kidney(L) & Kidney(R) & Liver & Pancreas & Spleen & Stomach\\
        \hline
            U-Net \cite{ronneberger2015u} & 76.85& 89.07 & 69.72 & 77.77 & 68.60 & 93.43 & 53.98 & 86.67 & 75.58\\
            TransUnet \cite{chen2021transunet} &77.48& 87.23 & 63.13 & 81.87 & 77.02 & 94.08 & 55.86 & 85.08 & 75.62\\
            Swin-Unet \cite{swinunet} &79.13& 85.47 & 66.53 & 83.28 & 79.61 & 94.29 & 56.58 & 90.66 & 76.60\\
            HiFormer \cite{heidari2023hiformer} &80.69& 87.03 & 68.61  &84.23 &78.37& 94.07 & 60.77  &90.44  & 82.03\\
        \hline
            GrabCut \cite{rother2004grabcut} &42.46 & 27.87 & 0.00 &67.46&59.87 & 72.39&3.42 &41.08&33.85  \\
            iSegFormer(2D)-p1 \cite{liu2022isegformer}& 74.62 & 82.88 & \textcolor{blue}{58.08} & 75.05 &72.23 &78.46 & 62.89 & 79.13 &64.67 \\
            iSegFormer(2D)-p3 \cite{liu2022isegformer}& 86.43& 89.07& 70.53& 89.60 & 86.92 &89.73 & 77.25 & 88.71 &83.22 \\
            iSegFormer(2D)-p5 \cite{liu2022isegformer}& 89.74& 91.43& 79.14& 91.56 &90.61 &92.20 & 82.42 & 90.57 &88.61  \\
            SAM-p1 \cite{kirillov2023segment}& 75.33 & \textcolor{blue}{88.98} & 52.74 & \textcolor{blue}{87.05} & \textcolor{blue}{85.98} & 72.41 & 41.19 & 78.06 & 64.44\\
            SAM-p3 \cite{kirillov2023segment}&78.61 & 87.93 & 52.22 & 86.76 & 86.04 & 81.59 & 48.27 & 79.29 & 71.21\\
            SAM-p5 \cite{kirillov2023segment}& 79.64 & 87.37 & 53.26 & 86.71 & 85.96 & 84.26 & 51.18 & 81.23 & 74.53\\
        \hline
            Ours-p1  & \textcolor{blue}{80.76}  & 82.67 & 57.05 & 86.46 & 85.85 & \textcolor{blue}{84.31} & \textcolor{blue}{67.97} & \textcolor{blue}{87.96} & \textcolor{blue}{73.86} \\
            Ours-p3  & \textcolor{blue}{90.51}  & \textcolor{blue}{89.43} & \textcolor{blue}{77.99} & \textcolor{blue}{91.57} & \textcolor{blue}{91.75} & \textcolor{blue}{94.53} & \textcolor{blue}{83.45} & \textcolor{blue}{94.54} & \textcolor{blue}{90.66} \\
            Ours-p5 & \textcolor{blue}{92.76}  & \textcolor{blue}{91.68} & \textcolor{blue}{84.81} & \textcolor{blue}{92.88} & \textcolor{blue}{92.88} &\textcolor{blue}{96.00} & \textcolor{blue}{87.52} & \textcolor{blue}{95.43} &\textcolor{blue}{94.15} \\
        \bottomrule 
        \end{tabular}
    }
    \label{tab:tab1}
\end{table}

\textbf{Materials and Implementation Details.} 
We validate our proposed framework on the public multi-organ dataset (named \textbf{Synapse}~\cite{synapse_dataset}) proposed in the 2015 MICCAI Multi-Atlas Abdomen Labeling Challenge. 
\textbf{Synapse} includes 30 cases with a total of 3779 axial abdomen 2D images where each CT volume involves 85-198 slices. 
The dataset is divided randomly into 24 cases for training, and 6 cases for testing. Eight organs are annotated by experts, including Aorta, Gallbladder, Left Kidney, Right Kidney, Liver, Pancreas, Spleen and Stomach.
To further test the performance of our proposed PE-MED, we built another dataset including different modalities (CT\&MRI) and 12 common Organs/Lesions, named \textbf{OL12}.
We randomly split \textbf{OL12} into 5050, 2041 and 4912 images for
training, validation and testing at the case level.

We implemented our framework using PyTorch using one NVIDIA 3090 GPU with 24 GB of memory. The input image sizes are $224\times224$ and $256\times256$ in \textbf{Synapse} and \textbf{OL12} datasets, respectively.
The models are trained for 100 epochs using a batch size of 128 and 64 for each dataset, using the normalized focal loss function~\cite{sofiiuk2022reviving}.
We optimized our model using the Adam optimizer, starting with a learning rate of $5 \times 10^{-3}$, and reducing the learning rate by a rate factor of 0.6 every 20 epochs. The dice score coefficient (DSC) was adopted to quantitatively evaluate segmentation performance. 
We also evaluated the methods using \textit{number of clicks} (NoC@\dag) metric, to measure the number of interactions required to achieve a predefined DSC (\dag).

\begin{table}[!t]
    \caption{Method comparison on \textbf{OL12} dataset, evaluated by DSC.}
     \centering
        \resizebox{\textwidth}{!}
    {
    \begin{tabular}{c|ccccc|cc}
    \hline
        Method & Point:1 & Point:2 & Point:3 & Point:5& Point:10 & NoC@85 & NoC@90  \\ 
        \toprule  
        iSegFormer(2D) \cite{liu2022isegformer}& 47.03(20.95) & 58.27(17.66) & 65.87(14.37) & 76.17(9.59) &85.83(4.89) &7.51(2.19) &9.56(1.13) \\ 
        SAM \cite{kirillov2023segment} & 51.07(34.85) & 56.90(32.96) & 58.49(32.79) & 61.31(32.23) &57.28(33.31) &6.83(4.12) &7.67(3.80)\\ 
        \hline
        Ours & \textcolor{blue}{86.49(12.25)} & \textcolor{blue}{91.35(7.41)} & \textcolor{blue}{92.91(5.65)} & \textcolor{blue}{94.24(3.97)} & \textcolor{blue}{95.18(2.97)} & \textcolor{blue}{1.65(1.45)} & \textcolor{blue}{2.39(2.37)} \\ 
        \bottomrule 
    \end{tabular}
    }
    \label{tab:tab2}
\end{table}

\textbf{Quantitative and Qualitative Analysis.} 
For \textbf{Synapse} dataset, we compared the proposed PE-MED with the fully supervised image segmentation methods, including TransUnet \cite{chen2021transunet}, Swin-Unet \cite{swinunet}, and HiFormer \cite{heidari2023hiformer}. Moreover, we also evaluated the performance of PE-MED with interactive segmentation methods such as GrabCut \cite{rother2004grabcut}, iSegFormer (2D version)~\cite{liu2022isegformer}, and SAM \cite{kirillov2023segment}.
For \textbf{OL12} dataset, we only compared PE-MED with two SOTA interactive segmentation methods, i.e., iSegFormer \cite{liu2022isegformer} and SAM \cite{kirillov2023segment}.
The ablation study was conducted on \textbf{Synapse} dataset, by comparing the PE-MED with different components, including Self-Loop (Baseline-SL), PALM-I (BaseLine-I), PALM-O (BaseLine-O), PALM (BaseLine-IO), and TSIP (BaseLine-T). 
Quantitative results are presented in Table \ref{tab:tab1}-\ref{tab:ablation}, with best results shown in \textcolor{blue}{Blue}.

For Table~\ref{tab:tab1}, it can be observed that our method achieves an average DSC of 80.76\% with a single point prompt, slightly higher than the current SOTA fully automated segmentation algorithm, i.e., HiFormer (80.69\%). 
With more prompts (\textit{-p5}), our method achieve a DSC of 92.76\%, outperforming all the reported fully automatic and interactive segmentation methods.
It is also notable that our proposed method shows the best performance compared to interactive methods under the same number of prompts.
Specifically, the DSC are 6.14\%/5.43\% (\textit{p1}), 4.08\%/11.90\% (\textit{p3}), 3.02\%/13.15\% (\textit{p5}) higher than iSegFormer and SAM, respectively.
Results on \textbf{OL12} dataset are reported in Table~\ref{tab:tab2}.
It can be seen that iSegFormer and SAM struggle to obtain satisfactory DSC even with 5 or 10 point prompts (\textless86\%).
However, our method achieved the DSC of 86.49\% even with one user interaction, and can outperform iSegFormer with 5\&10 prompts by about 20\%\&10\% DSC, respectively.
The last two columns (NoC@\dag) reveals that our method only require average 1.65 and 2.39 to reach a DSC of 85\% and 90\%, respectively.
While the other two methods require more user interaction to satisfy the corresponding requirements.

Results of ablation study can be found in Table \ref{tab:ablation}.
Experiments validated that by adding our proposed modules separately on the basis of baseline, the DSC performance can be improved.
Simultaneously, integrating all the modules (\textbf{Ours}) can further enhance the performance of the network.
It can also be observed that PE-MED achieves the highest DSC for six out of the eight organs, with only a slight deviation from the best results for the remaining two organs.
Specifically, for the most challenging organ, i.e., \textit{Gallbladder} with DSC of 74.17\% in Baseline, our proposed PE-MED can increase it by 10.64\%.
\begin{table}[!t]
    \caption{Ablation study under the setting of five point prompts.}
    \centering
    \resizebox{\textwidth}{!}
    {
        \begin{tabular}{c|c|cccccccccccccc}
        \toprule 
        \multirow{2}{*}{Methods}
        & \multicolumn{9}{c}{DSC(\%) $\uparrow$} \\
             \cline{2-10} 
            &AVG & Aorta & Gallbladder & Kidney(L) & Kidney(R) & Liver & Pancreas & Spleen & Stomach\\
        \midrule  
            Baseline &89.11& 90.25 & 74.17 & 89.14 &88.40 & 93.71 & 78.03 & 92.00 & 91.25 \\
            Baseline-SL & 91.70 & 93.12 & 73.02 & 91.77 & 91.92 & 95.13 & 83.79 & 94.45 & 93.11\\
            Baseline-I &91.84 & \textcolor{blue}{93.79} & 73.00 & \textcolor{blue}{93.07} & 92.10 & 95.40 & 82.69 & 94.93 & 91.70\\
            Baseline-O &91.22 & 92.29 & 71.63 & 91.84 & 91.91 & 95.06 & 83.12 & 93.62 & 92.56 \\
            Baseline-IO &92.24 &91.15 & 83.00 & 92.15 & 92.54 & 95.46 & 87.35 & 94.85 & 93.84 \\
            Baseline-T & 91.01 & 91.90 & 72.62 & 91.17 & 91.15 & 94.79 & 82.65 & 93.73 & 93.18\\
        \hline
            \textbf{Ours} & \textcolor{blue}{92.76}  & 91.68 & \textcolor{blue}{84.81} & 92.88 & \textcolor{blue}{92.88} &\textcolor{blue}{96.00} & \textcolor{blue}{87.52} & \textcolor{blue}{95.43} &\textcolor{blue}{94.15} \\
        \bottomrule 
        \end{tabular}
    }
    \label{tab:ablation}
\end{table}

\begin{figure}[!t]  
    \centering
    \includegraphics[width=1\textwidth]{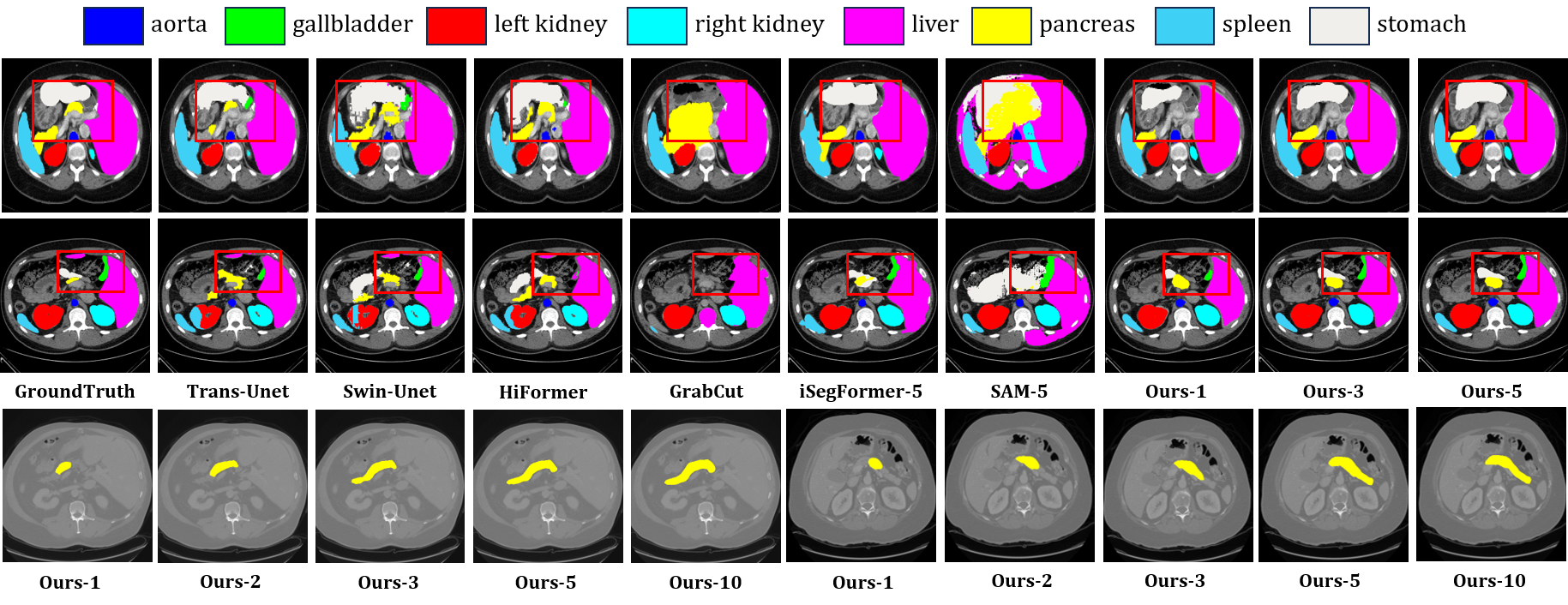}
    \caption{Segmentation performance on the \textbf{Synapse} and \textbf{OL12} dataset. The red rectangles highlight organ regions where the superiority of \textit{Ours} is evident.}
    \label{fig:vis_result}
\end{figure}

\begin{figure}[!t]  
    \centering
    \includegraphics[width=0.9\textwidth]{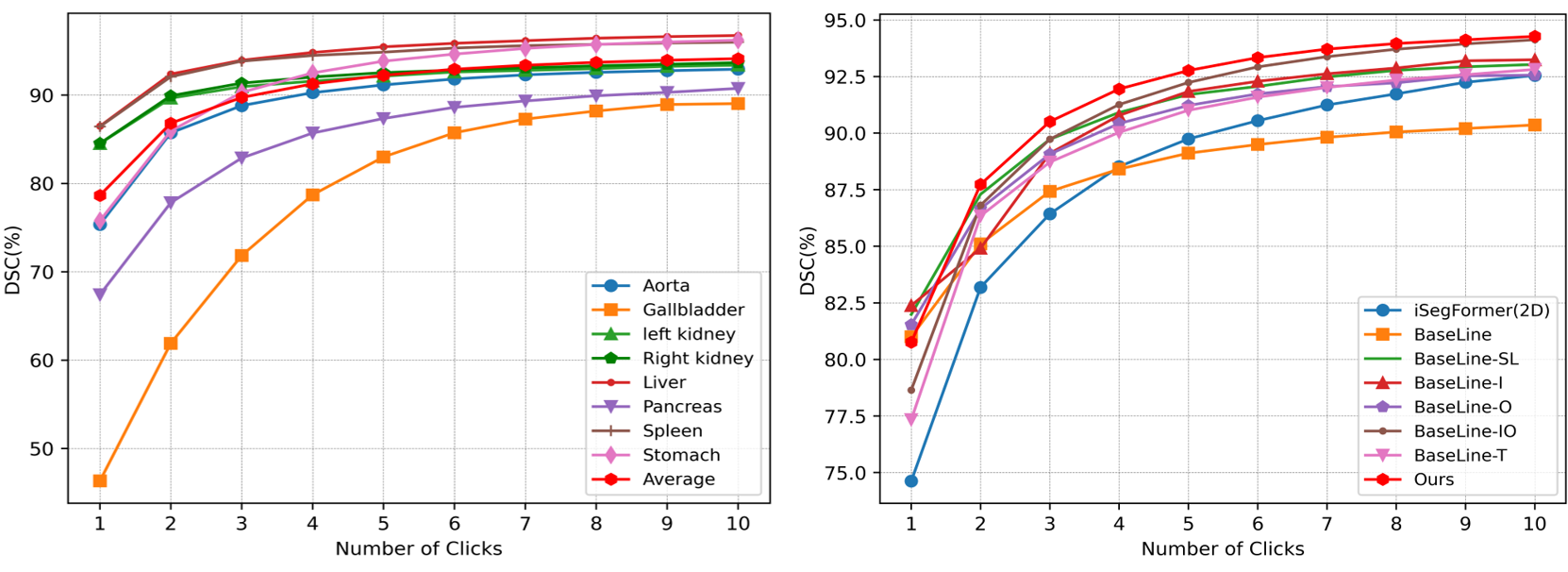}
    \caption{DSC variation tendency curve on \textbf{Synapse} dataset.}
    \label{fig:curve}
\end{figure}
 
Fig.~\ref{fig:vis_result} depicts the visualization results of different methods (rows 1-2), demonstrating our method as the closest to the Ground truth. Row 3 of Fig.~\ref{fig:vis_result} presents the visualization results for various numbers of interactions. Furthermore, increasing the number of interactions in our methodology leads to a progressive improvement in performance.
We also test the number of clicks and the results are shown in Fig.~\ref{fig:curve}.
In the left figure, we observe the rising trend of DSC for different organs and the average value, from clicks 1 to 10. Specifically, see the \textit{Average} curve, clicks 1-5 gain a higher improvement than clicks 6-10. Thus, we consider 5 clicks a suitable choice, since it is a trade-off between the NoC and DSC performance.
In the right figure, compared with \textit{Ours} and the ablation results, the contribution of our proposed prompt enhancement techniques can be validated.
Besides, it can be found that \textit{Ours} outperforms \textit{iSegFormer} at every click (red and blue curves), which further illustrates the power of our method.

\section{Conclusion}
In this work, we introduce an interactive framework for medical image segmentation, named PE-MED. 
Via the click user prompts, PE-MED can progressive optimize the segmentation results.
We proposed three techniques for enhancing the prompt information, including 1) Self-Loop strategy for providing warm initialization at the first interaction, 2) PALM for feature aggregation at one click, and 3) TSIP for temporal modeling among multiple interactions.
Extensive experiments on
two large datasets validate that PE-MED is general and efficient, achieving the best DSC results among all the strong competitors.
In the future, we will extend a 3D version PE-MED to directly handle the volumetric data.

\subsubsection{Acknowledgement.} 
This work was supported by the grant from National Natural Science Foundation of China (Nos.62171290, 62101343, 62101342), Shenzhen-Hong Kong Joint Research Program (No.SGDX20201103095613036), Shenzhen Science and Technology Innovations Committee (No.20200812143441001), and Guangdong Basic and Applied Basic Research Foundation (No.2023A1515012960).

\bibliographystyle{splncs04}
\bibliography{ref}

\end{document}